\documentclass{article}

\usepackage{multirow}
\usepackage{blindtext}
\usepackage{graphicx}
\usepackage{subcaption}
\usepackage{amsmath}
\usepackage{amssymb}
\usepackage{color}
\usepackage{soul}
\usepackage{fancyhdr}
\usepackage{pifont}
\newcommand{\cmark}{\ding{51}}%
\newcommand{\xmark}{\ding{55}}%
\definecolor{pink}{rgb}{0.858, 0.188, 0.478}

\begin{document}

\title{Denoising Adversarial Autoencoders: Classifying Skin Lesions Using Limited Labelled Training Data}

\author{Antonia Creswell\footnote{Corresponding Author: (ac2211@ic.ac.uk), \newline This paper is a preprint of a paper submitted to IET Computer Vision Journal. If accepted, the copy of record will be available at the IET Digital Library} , Alison Pouplin, Anil A Bharath}

\maketitle

\begin{abstract}
We propose a novel deep learning model for classifying medical images in the setting where there is a large amount of unlabelled medical data available, but labelled data is in limited supply. We consider the specific case of classifying skin lesions as either malignant or benign. In this setting, the proposed approach -- the semi-supervised, denoising adversarial autoencoder -- is able to utilise vast amounts of unlabelled data to learn a representation for skin lesions, and small amounts of labelled data to assign class labels based on the learned representation. We analyse the contributions of both the adversarial and denoising components of the model and find that the combination yields superior classification performance in the setting of limited labelled training data.
\end{abstract}

\section{Introduction}

The problem of image classification is one of assigning one or more labels to a given image. Deep learning has been demonstrated to be able to achieve both human and super-human levels of performance \cite{Esteva2017}, on classification tasks. However, achieving competitive levels of performance using deep learning often requires vast numbers of \{image, label\} pairs, typically in the millions. \\

In the medical image setting, it is unlikely that vast amounts of labelled images are available, particularly since medical experts are required to label the data, and this may be very costly and time consuming. Instead, it is often the case that there exists a large corpus of unlabelled data and a smaller dataset of labelled data. \\


We propose a model that is able to learn both from labelled data and from unlabelled data by building on previous work involving autoencoders \cite{bengio2013generalized, kingma2013auto, makhzani2015adversarial, vincent2008extracting, im2015denoising}. Autoencoders are able to learn data representations from unlabelled data, by jointly learning an encoder and decoder. The encoder maps data samples -- in this case images -- to a low dimensional encoding space, and the decoder maps the encoding back to image space. An autoencoder is trained to reconstruct its input. There are two key factors that enhance the performance of autoencoders, these are:

\begin{itemize}
    \item Denoising: Before being encoded, an input image is corrupted, and the decoder is trained to recover the clean image. By making the decoding process more challenging, the autoencoder learns more robust representations \cite{vincent2008extracting, vincent2010stacked}.\\

    \item Regularisation: Rather than allowing encoded data samples to occupy an unconstrained space, the distribution of encoded samples may be shaped to match a desired, \textit{prior} distribution, for example a multivariate standard normal distribution. Regularisation reduces the amount of information that may be held in the encoding, forcing the model to learn an efficient representation for the data.
\end{itemize}

To implement a denoising process, an arbitrary corruption process may be used. For example, white Gaussian noise \cite{bengio2013generalized} may be added to samples of the training data. Corruption is often trivial to implement. More challenging is the regularisation of the distribution of encoded data samples. There are at least two approaches for shaping the distribution of encoded samples to match a desired distribution. The two key methods for regularising the encoding space are:

\paragraph{\bf Variational} Minimising the KL divergence between the distribution of encoded samples and a chosen prior distribution \cite{kingma2013auto}. For ease of implementation, the prior distribution is often a multivariate standard normal distribution and the encoder is designed to learn parameters of a Gaussian distribution.

\paragraph{\bf Adversarial} Rather than using the encoder to parametrise a distribution and calculate the KL divergence, a third, discriminative model is trained to correctly distinguish encoded samples from samples drawn from a chosen prior distribution. The encoder is then updated to encode samples such that the discriminator cannot distinguish encoded data samples from samples drawn from the prior distribution \cite{makhzani2015adversarial}. We will more formally introduce adversarial training in Section \ref{sec:AAE}.\\

The adversarial \cite{makhzani2015adversarial} approach allows the encoder to be more expressive than the variational approach \cite{kingma2013auto}, and has achieved superior classification performance in a semi-supervised fashion on several benchmark dataset. While denoising and adversarial training have been used to augment autoencoders in isolation, they have yet to be combined in one model. Here, we propose augmenting an autoencoder with both a denoising criterion and by using adversarial training to shape the distribution of encoded data samples. We augment this model further to make use of labelled data where it is available while still learning from unlabelled data where label information is not available.\\

Our contributions are as follows:

\begin{itemize}
    \item We introduce the semi-supervised denoising adversarial autoencoder (ssDAAE) which is able to learn from a combination of labelled and unlabelled data (Section \ref{sec:SSDAAE}).\\

    \item We apply our model, the ssDAAE, to the task of classifying skin-lesions as benign or malignant in the setting where the amount of labelled data is limited (Section \ref{sec:method}).\\

    \item We compare performance of the ssDAAE with a semi-supervised adversarial autoencoder (ssAAE), a fully supervised AAE (sAAE), a fully supervised DAAE (sDAAE), and a CNN trained with and without corruption. For fair comparison, the CNNs had the same architecture as the encoder of the ssAAE and ssDAAE; that is, the portion of the ssAAE and ssDAAE architecture used to perform classification is the same as the CNN used for standard deep network classification. Additionally, we assessed the effect of additive noise during training of the otherwise standard CNN. Our results show that the ssDAAE consistently out performs the others.\\

\end{itemize}

Although we demonstrate this approach on skin lesions, the semi-supervised approach explored in this paper are not specific to skin lesions, and could potentially be applied to other image datasets where labelled samples are in limited supply, but there is a surplus of unlabelled images that have been captured. 

\section{Method: Classifying Skin Lesions}

In this section, we formulate the ssDAAE. First, we discuss the skin lesion classification problem. Secondly, we describe the Adversarial Autoencoder (AAE) and then we describe how the AAE may be augmented to become the ssDAAE. Finally, we describe how the ssDAAE is trained.

\subsection{Skin lesion classification}

\begin{figure}[h]
 \centering
\begin{subfigure}{0.45\columnwidth}
\includegraphics[width=0.9\linewidth]{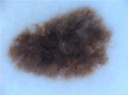} 
\caption{Benign}
\end{subfigure}
\begin{subfigure}{0.45\columnwidth}
\includegraphics[width=0.9\linewidth]{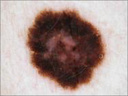}
\caption{Malignant}
\end{subfigure}
 
\caption{\textbf{Examples of Benign and Malignant skin-lesions.} Classifying skin lesions as benign or malignant is non-trivial and requires expert knowledge.}
\label{fig:skin_lesions}
\end{figure}

Skin lesion classification is a non-trivial problem. Even humans have to be specially trained to be able to distinguish benign (not harmful) skin lesions from malignant (harmful) skin lesions. Examples of benign and malignant skin lesions are shown in Figure \ref{fig:skin_lesions}. The high level goal is to train a model to correctly predict whether a skin lesion is benign or malignant. Beyond this, we want to design models, for which we can be confident that we correctly identify a specific proportion of malignant skin lesions as being malignant, while still being able to correctly identify a large number of benign skin lesions as being benign. To this end, in the following sections we describe the model that we propose for skin lesion classification in the setting of limited labelled data.

\subsection{Adversarial Autoencoders}
\label{sec:AAE}

An autoencoder consists of two models, and encoder and a decoder, each with their own set of learnable parameters. In our approach, we are using deep convolutional neural networks to embody the encoder and decoder. The encoder, $ E_{\theta_E} :x \rightarrow \hat{z}$  with parameters $\theta_E$, is designed to map an image sample, $x$ to an encoding, $\hat{z}$. The encoding vector, $\hat{z}$, is of much lower dimension than the number of pixels in an image, $x$. The decoder, $D_{\theta_D}: \hat{z} \rightarrow \hat{x}$ is designed to map an encoding $\hat{z}$ back to an image, $\hat{x}$. The parameters, $\theta_E$ and $\theta_D$ of the encoder and decoder respectively are learned such that the  difference between the input to the encoder, $x$, and the output of the decoder, $\hat{x}$, are minimised.\\ 

The adversarial autoencoder \cite{makhzani2015adversarial} incorporates adversarial training \cite{goodfellow2014generative} to shape the distribution of encoded data samples to match some chosen prior distribution, $p(z)$, such as a multivariate standard normal distribution. Note that we are applying adversarial training to the encoded data samples, rather than the data samples, as more commonly seen in the literature \cite{goodfellow2014generative, radford2015unsupervised}. Adversarial training requires the introduction of another model, a discriminator, for which we also use a deep convolutional neural network. The discriminator, $T_{\theta_T}: z \rightarrow (0,1)$ maps encodings (either encoded data samples, $\hat{z}$ or samples drawn from the prior, $z$,) to a probability of whether that sample comes from the chosen prior distribution. The parameters, $\theta_T$ of the discriminator are learned such that high values are assigned to samples that come from the chosen prior distribution and low values are assigned to samples that come from the encoder. To encourage encoded samples to match the chosen prior distribution, the parameters of the encoder, $\theta_E$ are updated such that $T_{\theta_T}(E_{\theta_E}(x))$ are maximised.\\

Formally, the following objectives must be optimised during the training of an adversarial autoencoder:

\begin{equation}
\label{recCost}
    \arg \min_{\theta_E, \theta_D} || x - D_{\theta_D}(E_{\theta_E}(x)) ||_2^2 
\end{equation}


{\begin{equation}
\label{disCost_z}
    \arg \max_{\theta_T} \mathbb{E}_{z \sim p(z)} \log T_{\theta_T}(z) + \mathbb{E}_{\hat{z} \sim p_{g(z)}} \log (1 - T_{\theta_T}(\hat{z})) 
\end{equation}
where $p_g(z)$ is the distributions of encoded data samples, $\hat{z}=E_{\theta_E}(x)$ and $x \sim \mathcal{D}$.}

\begin{equation}
\label{encCost}
    \arg \min_{\theta_E} \mathbb{E}_{x \sim \mathcal{D}(x)} \log T_{\theta_T}(E_{\theta_E}(x))  
\end{equation}

where $p(z)$ is some chosen prior distribution, for example a standard normal and $\mathcal{D}(x)$ is the training data distribution.\\

Equation (\ref{recCost}) is the reconstruction cost that is used to train the encoder and decoder. This cost should be minimised, so that input images may be recovered after encoding and decoding. Equation (\ref{disCost_z}) is the discriminator cost, which, when minimised, means that the discriminator can correctly distinguish between encoded data samples and samples from the chosen prior distribution. Equation (\ref{encCost}) is a second regularisation cost used to update the encoder. When minimised -- simultaneously with Equation (\ref{encCost}) \cite{goodfellow2014generative} -- this regularisation cost encourages the distribution of encoded samples to be similar to the chosen prior distribution.\\

An adversarial autoencoder may be trained entirely with unlabelled data and may be evaluated by measuring reconstruction error on a test dataset, or synthesising novel samples by first drawing samples, $z$, from the chosen prior distribution and passing these through the decoder to produce synthetic images, $\hat{x}$. The process of encoding and decoding test samples often reveals whether or not the decoder model has learned a sufficient representation for the data. A further test is to attempt to generate novel samples, by passing random encodings -- drawn from the chosen prior distribution -- through the decoder. Since a regularised autoencoder is able to generate novel samples, we often refer to it incorporating a {\em generative model} of the data.\\

In its current form, it is not immediately obvious how an adversarial autoencoder may be used to perform classification. In fact, it is necessary to augment the encoder to predict not only the encoding, but also the label.


\subsection{Semi-Supervised Denoising Adversarial Autoencoder}
\label{sec:SSDAAE}

Before learning to classify skin lesion as benign or malignant, we may first consider learning more about what skin lesion looks like. This could involve learning the colour, general shape and textures of skin lesions. An ssDAAE allows us to do this by incorporating both a generative and classification model in one. The ssDAAE differs from the AAE in two ways.\\

Firstly, the AAE is augmented by applying a corruption process. The corruption process, $C : x \rightarrow \tilde{x}$ is a stochastic process in which Gaussian noise, with standard deviation, $\sigma$, is added to a data sample, $x$ to obtain, $\tilde{x}$. This change results in a DAAE.\\

Secondly, the encoder of the DAAE is altered to define an ssDAAE by splitting the encoder in to three parts. An initial encoder, $E_{\theta_E}: \tilde{x} \rightarrow \hat{h}$, and two sub-encoders, $E^y_{\theta_y}: \hat{h} \rightarrow \hat{y}$ and $E^z_{\theta_z}: \hat{h} \rightarrow \hat{z}$. The encoder is trained to predict not only an encoding, $\hat{z}$ but also a label vector, $\hat{y} \in (0,1)$. Adversarial training is used (as in an AAE \cite{makhzani2015adversarial}) to shape both the distribution of encoded samples to match a chosen prior distribution, and the distribution of predicted class labels to match a categorical distribution \cite{makhzani2015adversarial}.\\

However, since we are posing skin lesion classification as a binary classification problem, we represent the labels benign and malignant using a single unit and apply a sigmoid function at the end of $E^y_{\theta_y}$. We therefore train a label discriminator $T^y_{\theta_{T_y}}$ to distinguish predicted labels $\hat{y}$ from labels drawn from a binary distribution. This encourages the output of the classifier, $E^y_{\theta_y}$, to be either $0$ or $1$ rather than taking values in between.\\




For an input $x$, the output of the decoder, $\hat{x}$ is given by:
\begin{equation}
    \hat{h}=E_{\theta_E}(C(x))
\end{equation}

\begin{equation}
    \hat{x} = D_{\theta_D}([E^y_{\theta_y}(\hat{h}),E^z_{\theta_z}(\hat{h}) ])
\end{equation}
where, $[a,b]$ is a concatenation of $a$ and $b$.\\

The weights of the encoder, $\theta_E$ and $\theta_z$, are updated via both adversarial training to match the distribution of $\theta_z$'s to a chosen prior and a reconstruction error between $x$ and $\hat{x}$. This forms the generative part of the model, and may be trained on entirely unlabelled data. This property of the model means that we can learn parameters $\theta_E$ and $\theta_z$ using large amounts of unlabelled data to learn more about the structure of skin-lesions. We can also visualise what this model has learned by generating novel images of skin lesions and evaluating them by eye to see whether the model has captured the  basic concept of what a skin lesion is.\\

Following on from this, we may use the limited labelled training data to ``fine tune'' the generative model. We may use the labelled data to update the weights $\theta_y$, or additionally to update $\theta_E$ by minimising the classification error between predicted a label $\hat{y}$ and the true label $y$. Experimentally, we found it beneficial to update both $\theta_y$ and $\theta_E$ as this made training more stable.\\

\begin{figure}[ht]
    \centering
    \includegraphics[width=0.99\columnwidth]{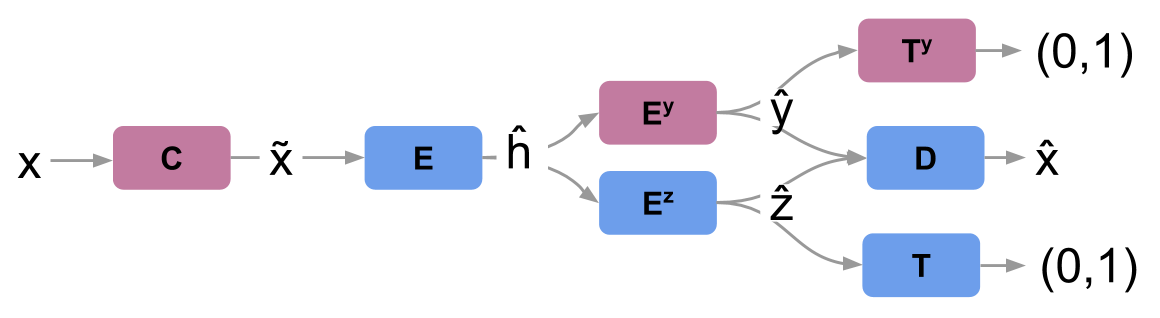}
    \caption{\textbf{ssDAAE model} Image data, $x$, is corrupted before being encoded. The encoding process consists of two sub-mappings of the corrupted image, $\tilde{x}$, yielding an encoding of the image appearance $\hat{z}$ and a label prediction, $\hat{y}$. The decoder uses both of these to reconstruct a version of the uncorrupted image, $\hat{x}$. The blue parts correspond to an AAE model, while the red parts are additions that make this model an ssDAAE.}
    \label{fig:ssDAAE}
\end{figure}

For completeness, note that -- similar to an adversarial autoencoder (AAE) -- the weights of the decoder, $\theta_D$ are learned as part of the minimisation of the reconstruction error between $x$ and $\hat{x}$. In Figure \ref{fig:ssDAAE}, we present a diagram of our proposed model. 

\subsection{Training Data}
\label{sec:dataset}

As described above our ssDAAE may be trained using a mixture of both labelled and unlabelled data. The labelled data is obtained from the ISIC archive \cite{ISIC}. The archive consists of nearly $14$k images, of which over $9$k are benign skin lesions taken from children. The child skin lesion samples contain colour-coded identifier patches - rendering them not suitable for training in their current state. Of the remaining images there are $3419$ examples of benign skin lesions and $1082$ examples of malignant skin lesions. \\



\begin{figure}
    \centering
    \includegraphics[width=0.9\columnwidth]{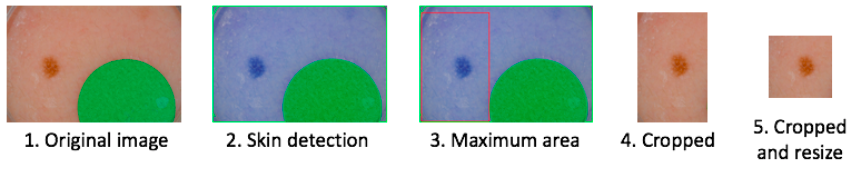}
    \caption{\textbf{Pre-processing of child skin lesions}  Images have been pre-processed to crop out the area of the skin lesion. First, the skin was detected using a color profile that matches any usual skin color within a certain threshold in order to obtain a binary mask, in which a rectangle was defined such that it possesses the maximal area. The image was then cropped and centered in order to obtain square images of 64x64.}
    \label{fig:preproc}
\end{figure}

To make the $9$k skin lesions taken from children more appropriate for training and classification, we removed the identifier patches as shown in Figure \ref{fig:preproc}. This processing step is not considered to be part of the classification framework, rather a means to increase the amount of available training data. These identifier patches are unlikely to be present in real world encounters. The processed child skin lesions are combined with the rest of the benign skin-lesions.\\


The ISIC dataset \cite{ISIC} does not specifically provide distinct labelled and unlabelled datasets. We partition the data into $7$k unlabelled data samples, $5$k labelled data samples to be used for training and the rest, $500$ each for testing and validation. To expand each dataset we performed data augmentation, by flipping the examples of skin-lesions in both the x and y axes and rotating the samples up to $180$ degrees.

\section{Experiments and Results}
\label{sec:method}

In this section, we perform exhaustive ablation study to isolate the effectiveness of (a) a model that incorporates denoising, (b) the use of an adversarial autoencoder (generative model) opposed to a CNN (discriminative model) and (c) of utilising additional unlabelled data. From these experiments we will be able to isolate exactly which components of the ssDAAE are necessary to achieve good performance. We start by explaining how the performance of our models is evaluated.

\subsection{Evaluation}
\label{sec:eval}
There is significant label imbalance that can be observed in the ISIC dataset, meaning that the majority of the images ($c.90$\%) are benign, choosing a single classification accuracy as a performance metric may have been misleading given that even a system that always outputs the benign class would get, on average, a high score ($c. 0.90$). Instead, we prefer using clinically insightful and interpretable metrics such that the percentage of malignant skin lesions correctly classified as malignant (true positive, or sensitivity) and benign skin lesions correctly classified as benign (true negative, or specificity). Furthermore, in the context of a medical application and because of the label imbalance problem, we are particularly interested in comparing the model performances, in terms of specificity, at high sensitivity values, to avoid miss diagnosing a malignant skin lesion as benign. \\


We used similar evaluation metrics to those using in an interrelation skin lesion classification challenge, hosted by the International Skin Imaging Collaboration (ISIC), at the International Symposium on Biomedical Imaging (ISBI). The challenge was composed of three tasks, where the final tasks was skin lesion classification. For the last task, participant's models were ranked according to the specificity of their model giving a particular sensitivity threshold $\{0.82, 0.89, 0.95, 0.99\}$. We use the same evaluation metrics in our in all our experiments.



\subsection{Training and Architectural Details}

\subsubsection{Architectural Details} 
\label{sub:architectural_details}

  In this subsection we present the detailed architecture of both the CNN baseline
  as well as our semi-supervised denoising adversarial autoencoder.

  \paragraph{CNN}
  The baseline CNN model consists of a sequence of 4 convolution layers, a relu non-linearity is applied to the output of each layer before being fed to the next one. The output of the CNN sequence is then flattened and fed to a linear layer containing $1000$ neurons followed by a final linear layer, with one neuron and sigmoid non-linearity that returns values between $(0,1)$, where $0$ is the label for benign and $1$ is the label for malignant.

  \paragraph{Semi-supervised Denoising Adversarial Autoencoder}

  The CNN described above, without the final linear layer, forms the encoder, $E_{\theta_E}$ of our adversarial autoencoder. The output of the 1000 neurons linear layer splits in two:
  
  \begin{enumerate}
  	\item \textbf{Latent Encoder, $E^z_{\theta_z}$} : consists of a linear layer with $200$ neurons that is responsible for returning the 200-dimension encoding vector, $\hat{z}$, representing an  input image in the learned latent space.
  	\item \textbf{Classifier, $E^y_{\theta_y}$} : consists of a linear layer with a single neuron followed by a sigmoid activation function. The output of this layer is the class prediction, $\hat{y}$, for a given input image.
  \end{enumerate}


  The label output $\hat{y}$ as well as the encoded vector $\hat{z}$ are then fed through three sub-networks (refer to Figure \ref{fig:ssDAAE} for a visualisation).

  \begin{itemize}
      \item \textbf{Latent Space Discriminator, $T_{\theta_T}$} : This model consists of a linear layer with $1000$ neurons and a relu non-linearity, followed by a linear layer with a single neuron and sigmoid non-linearity.
      \item \textbf{Binary Label Discriminator, $T^y_{\theta_{T_y}}$} : This model has similar architecture to the the Latent Space Discriminator.
      \item \textbf{Decoder, $D_{\theta_D}$} : Finally, this model consists of a linear layer followed by sequence of 4 transposed convolution layers, again a relu non-linearity is applied to the output of each transposed convolution layer before being fed to the next one, finally a sigmoid layer is applied to the last output of the sequence. The input to the decoder is the concatenation of the label and the encoded vector.
  \end{itemize}
 These models are summarised in the Appendix in Table \ref{CNN_architecture}.

\subsubsection{Preprocessing and Input Corruption}

All images have been scaled so that all values are between 0 and 1. Furthermore, In order to allow for the partial corruption of the input. A corruption layer has been implemented that is responsible for adding Gaussian noise with mean $0$ and variance $\sigma$ to the input of the system. Various $\sigma$ values between $0$ and $1$ have been attempted. Experimentally the best results were obtained for $\sigma = 0.1$.

\subsubsection{Loss functions and Class Imbalance}

In this section we describe in detail how we balance the cost functions used to train our networks. For training the baseline CNN model, a single binary cross entropy loss function was used as it is an appropriate loss function for classification tasks. The ssDAAE on the other hand consists of several modules each with their own loss function, these loss functions need to be combined with care. The loss functions include: a classification loss at the output of the classifier, $E^y_{\theta_y}$, a reconstruction loss at the output of the decoder, $D_{\theta_D}$ and both discriminator and regularisation losses at the output of the discriminators, $T_{\theta_T}$ and $T^y_{\theta_{T_y}}$. The latent space discriminator loss is described in Equation (\ref{disCost_z}), this cost may be modified for the label discriminator, by replacing $p(z)$ with a binary distribution and $p_g(z)$ with the output of $E^y_{\theta_y}$. We now describe the encoder loss function, (designed to update the weights, $\{\theta_E, \theta_{E_y}, \theta_{E_z} \}$), which is defined as the weighted combination of the following losses:\\


\begin{itemize}
    \item \textbf{Classification Loss $l_{class}$ : } The binary cross-entropy loss between the predicted class and the ground truth label.
    \item \textbf{Reconstruction Loss $l_{rec}$ : } The mean squared-error between the decoded image and the input image.
    \item \textbf{Latent Regularisation Loss $l_{reg_z}$ : } The binary-cross entropy loss between output of the latent discriminator, $T_{\theta_T}$ and a target label 1. (Where $1$ refers to the discriminator predicting that a samples is from the chosen prior distribution).
    \item \textbf{Label Regularisation Loss $l_{reg_y}$ : } The binary-cross entropy loss between output of the binary label discriminator, $T^y_{\theta_{T_y}}$ and a target label $1$. (Where $1$ refers to the discriminator predicting that a sample is from a binary distribution).
\end{itemize}

\begin{center}
$l_{encoder} = \beta l_{class} + \eta l_{rec} + \alpha (l_{reg_y} +  l_{reg_z})$
\end{center}

where $\alpha$, $\beta$ and $\eta$ are coefficients chosen through experimentation.\\

Furthermore, due to the heavy class imbalance in the ISIC dataset (90\% of the data is benign), it was also necessary to slightly modify the cross entropy loss function for the classification loss by adding a weight $a$ for label 1 and a different weight $b$ for label 0. which leads to the following expression :
\begin{center}
$l_{class} = a * (y * log(\hat{y})) + b * ((1-y) log(1-\hat{y}))$
\end{center}

\subsubsection{Hyper Parameter Choices}

For both the baseline model and our adversarial auto-encoder model, we used $(a,b) = (9,1)$ for the weighted classification loss. The CNN was trained using an RMSProp Optimizer with a momentum of $0$ and a learning rate of $10^{-4}$. The encoder and decoder of the ssDAAE were trained with the same optimizer with same learning are and momentum as the CNN. We found setting coefficients $(\alpha, \beta, \eta) = (0.1, 1, 0.1)$ to work well. When training the discriminator training, the same optimizer and learning rate were used, but the momentum was set to $0.2$.

\section{Ablation Study}
To appreciate the contributions of our proposed model, we performed ablation studies. We trained $6$ different models listed in Table \ref{table:ablation}. Each autoencoding model -- consisting of an encoder an decoder -- had the same architecture and each CNN had the same architecture as the encoder. The CNN and CNN+noise models act as simple baselines that do not incorporate a generative model, and are trained in a fully supervised way, not making use of any unlabelled data. The sAAE and sDAAE are fully supervised models, that do incorporate a generative model, in the form of an adversarial autoencoder. Finally, the ssAAE and ssDAAE are trained in a semi-supervised fashion to use both labelled and unlabelled data. All models were trained with the same amount of labelled data. The semi-supervised models are trained with the same amount of unlabelled data. To make the comparisons as fair as possible, we used the same hyper parameters \footnote{learning rate, number of training epochs, amount of labelled and unlabelled data, loss function weightings, level of corruption, size of encoding} for all models in the study.\\

\begin{table}[h]
\caption{Models used for the ablation study. The semi-supervised DAAE (ssDAAE) has three core components (a) denoising, (b) an adversarial autoencoder and (c) is trained in a semi-supervised fashion, training with additional unlabelled data. The sAAE and sDAAE are fully supervised models.\label{table:ablation}}
{\begin{tabular}{ c c c c c }
 \textbf{Model} & (a) \textbf{Denoising} & (b) \textbf{Autoencoder} & (c) \textbf{Unlabelled} \\
 \hline
 CNN & \xmark & \xmark & \xmark \\  
 CNN + noise & \cmark & \xmark & \xmark \\
 sAAE & \xmark & \cmark & \xmark \\    
 sDAAE & \cmark & \cmark & \xmark \\
 ssAAE & \xmark & \cmark & \cmark \\
 ssDAAE & \cmark & \cmark & \cmark \\
 \hline
\end{tabular}}{}
\end{table}

\begin{figure}
    \centering
    \includegraphics[width=0.99\columnwidth]{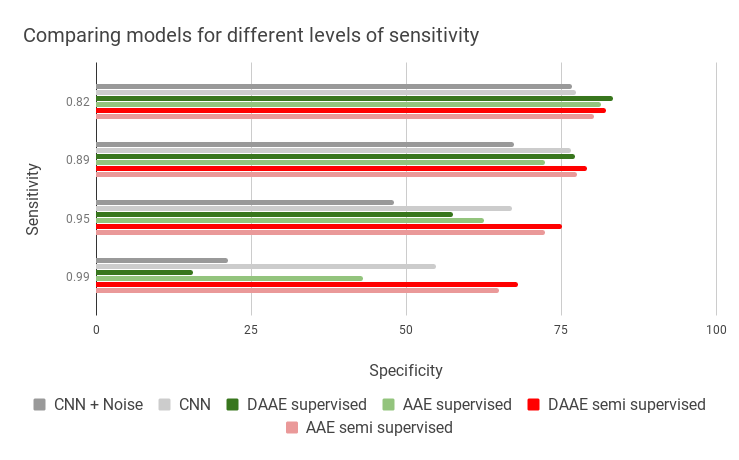}
    \caption{\textbf{Ablation Study on the ICIS image database.} The results of this study allow us to compare effect of different model variants. The ssDAAE yields the best specificity at high sensitivity levels.}
    \label{fig:ablation}
\end{figure}

The results of our ablation study are shown in Figure \ref{fig:ablation}. At all sensitivity values the ssDAAE outperformed the simple baselines, of the CNN and the CNN with added noise (CNN+noise). At all sensitivity values, the ssDAAE outperformed the ssAAE, suggesting that the corruption process is useful, but perhaps, more so in the semi-supervised model where there are more examples, since the sAAE outperformed the sDAAE only at lower sensitivities $(0.82, 0.89)$. \\

Additionally, the CNN outperformed the CNN+noise model at all sensitivity values, further suggesting that many more training examples are needed for denoising to be effective. The fact that the CNN+noise performed less well than a CNN for all sensitivities, in contrast to the sAAE and sDAAE, which do perform well at the lower sensitivities, may be because the CNN+noise network is never exposed to the uncorrupted images, while the autoencoder models are exposed to uncorrupted images when the reconstruction loss is computed. \\

It is at the higher sensitivities $(0.89, 0.95, 0.99)$ that we most clearly see that all semi-supervised variants outperformed their supervised variants. The benefits of semi-supervised models over fully supervised suggested by the results, supports our motivation to design models that incorporate unlabelled data with labelled. Further, the additional benefit of incorporating a denoising criterion into semi-supervised models has, as anticipated, also improved performance. Finally, our results suggest that the model that most consistently performs well is our proposed model, the ssDAAE.\\



\subsection{Effect of different levels of corruption}
\begin{figure}
    \centering
    \includegraphics[width=0.99\columnwidth]{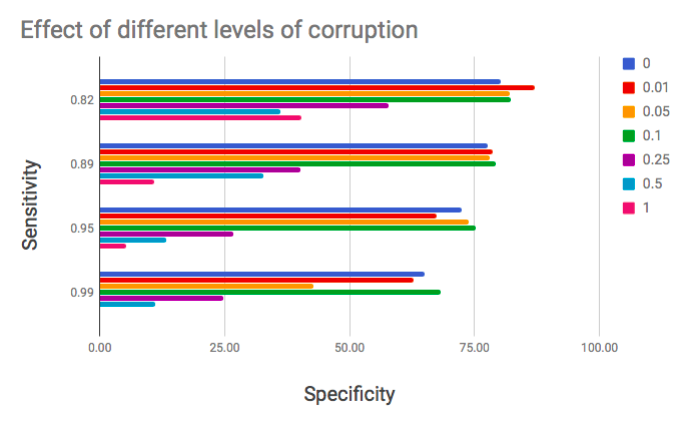}
    \caption{\textbf{Effect of the level of corruption when training an ssDAAE.} We compare models trained with corruption levels, $\sigma=\{0,0.1,0.25,0.5,1\}$. For lesion classification, moderate levels of noise yield the best results.}
    \label{fig:noise_levels}
\end{figure}

We also explored the effects of using different levels of corruption during training of ssDAAE models. We compared models trained using noise levels, $\sigma=\{0, 0.01, 0.05, 0.1, 0.25, 0.5, 1.0\}$, the results are shown in Figure \ref{fig:noise_levels}. Each model had the same architecture and was trained with the same hyper parameters \footnote{learning rate, number of training epochs, amount of labelled and unlabelled data, loss function weightings, size of encoding} to make the comparison as fair as possible. Our results suggest that the optimal corruption level is $\sigma=0.1$ for most sensitivity values. We see that, for all sensitivity values, an ssDAAE trained with a noise level of $\sigma=0.1$ outperformed an ssAAE (a model trained with a noise level of $\sigma=0$). For ssDAAE models trained with noise levels greater than $\sigma=0.25$, inclusive, performance dropped significantly for all sensitivity values, suggesting that too much noise may have an adverse effect on training.

\section{Related Work}

Previous research has been conducted on the ISIC dataset \cite{ISIC}, as part of a challenge hosted by ISBI \cite{ISBI}, however there are two core differences between our work and the approaches currently taken for this and other medical image datasets.\\

Firstly, while our work focuses on a single end-to-end classification approach, previous work on skin lesion classification has tended to adopt a three-stage approach \cite{codella2017skin, li2017skin, ramlakhan2011mobile}, splitting the task into, (1) a lesion segmentation to extract the relevant parts of the images \cite{schmid1997colour, denton1995boundary}, followed by (2) a dermoscopic feature classification \cite{kasmi2016classification, round2000lesion} that helps to detect clinical patterns, and, finally, (3) a disease classification task aiming to identify ``melanoma'', ``seborrheic keratosis'' and ``benign nevi''. The initial preprocessing stages, (1) and (2) require extensive pixel-wise labelling of images -- such as by image segmentation -- to provide ground truth examples in order to learn to perform these tasks. On the contrary, our approach requires only a small amount of labelled data, where the label is simply ``benign'' or ``malignant'', and makes use of unlabelled data, too.\\

The best performances recorded during this challenge have been obtained using fully-supervised deep learning architectures (AlexNet \cite{ISIC-Berseth}), and transfer learning (VGG-16 nets \cite{ISIC-Menegola17}, ResNet \cite{ISIC-BiLei}) with networks previously trained on ImageNet. While the success of approaches introduced by Menegola et al. \cite{ISIC-Menegola17} and Lei et al. \cite{ISIC-BiLei} highlight the benefits of using additional data to improve performance, the additional data they use is different to the skin lesion data. This because these approaches, \cite{ISIC-Menegola17}, \cite{ISIC-BiLei} are fully supervised and therefore can only make use of labelled data, the authors were therefore limited to use datasets for which labelled data was available. \\

One of our main contributions is in proposing a specific architecture and approach to skin lesion classification which can make use of both labelled and unlabelled medical image data. This method allows classification models to make use of unlabelled data, when the amount of labelled data is in limited supply. This allows us to use additional data that is more similar to the skin lesion data when training our models, unlike Menegola et al. \cite{ISIC-Menegola17} and Lei et al. \cite{ISIC-BiLei}.




\section{Conclusion}
Despite the clear success of deep learning techniques in specific image datasets, wide adoption of the many available approaches to training deep networks techniques is highly dependent on the availability of sufficient quantities of $\{label, image\}$ pairs. \\

The solution that we propose in this work is a form of semi-supervised learning, in the sense that if ground truth labels are available for only a subset of the data, all the data can still be used to train a deep classification model. Our results show that the additional information that may be learned from the unlabelled data is useful for boosting classification performance. \\

Our solution also includes a denoising procedure. While an adversarial autoencoder \cite{makhzani2015adversarial} is trained to simply to recover its input, our model is trained to recover clean data samples from corrupted ones. This results in our model learning a more robust data representation, which in turn boosts classification performance. \\

The approach we suggest is not limited specifically to the form of image data explored in this paper.  Currently, we have applied this to dermatological images of skin lesions. Our model is flexible, and may potentially be applied to other datasets, where there is a large amount of image data, but a limited amount of it is labelled. The semi-supervised approach that we have taken in this paper holds significant relevance in developing high specificity classification systems for other medical images. This is because it is often the case that it is very easy to collect many examples of unlabelled images and the availability of experts that provide ground truth labelling is limited. 



\section*{Acknowledgment}
We like to acknowledge the Engineering and Physical Sciences Research Council for funding through a Doctoral Training studentship as well as Nick Pawlowski and Martin Rajchl for help with providing access to cluster computers.

\bibliographystyle{abbrv}
\bibliography{bib}

\begin{thebibliography}{10}

\bibitem{ISIC}
International skin imaging collaboration: Melanoma project.

\bibitem{bengio2013generalized}
Y.~Bengio, L.~Yao, G.~Alain, and P.~Vincent.
\newblock Generalized denoising auto-encoders as generative models.
\newblock In {\em Advances in Neural Information Processing Systems}, pages
  899--907, 2013.

\bibitem{ISIC-Berseth}
M.~Berseth.
\newblock {ISIC} 2017 - skin lesion analysis towards melanoma detection.
\newblock {\em CoRR}, abs/1703.00523, 2017.

\bibitem{ISIC-BiLei}
L.~Bi, J.~Kim, E.~Ahn, and D.~Feng.
\newblock Automatic skin lesion analysis using large-scale dermoscopy images
  and deep residual networks.
\newblock {\em CoRR}, abs/1703.04197, 2017.

\bibitem{codella2017skin}
N.~C. Codella, D.~Gutman, M.~E. Celebi, B.~Helba, M.~A. Marchetti, S.~W. Dusza,
  A.~Kalloo, K.~Liopyris, N.~Mishra, H.~Kittler, et~al.
\newblock Skin lesion analysis toward melanoma detection: A challenge at the
  2017 international symposium on biomedical imaging (isbi), hosted by the
  international skin imaging collaboration (isic).
\newblock {\em arXiv preprint arXiv:1710.05006}, 2017.

\bibitem{ISBI}
N.~C.~F. Codella, D.~Gutman, M.~E. Celebi, B.~Helba, M.~A. Marchetti, S.~W.
  Dusza, A.~Kalloo, K.~Liopyris, N.~K. Mishra, H.~Kittler, and A.~Halpern.
\newblock Skin lesion analysis toward melanoma detection: {A} challenge at the
  2017 international symposium on biomedical imaging (isbi), hosted by the
  international skin imaging collaboration {(ISIC)}.
\newblock {\em CoRR}, abs/1710.05006, 2017.

\bibitem{denton1995boundary}
W.~Denton, A.~Duller, and P.~Fish.
\newblock Boundary detection for skin lesions: an edge focusing algorithm.
\newblock 1995.

\bibitem{Esteva2017}
A.~Esteva, B.~Kuprel, R.~A. Novoa, J.~Ko, S.~M. Swetter, H.~M. Blau, and
  S.~Thrun.
\newblock {Dermatologist-level classification of skin cancer with deep neural
  networks}.
\newblock {\em Nature}, 542(7639):115--118, jan 2017.

\bibitem{goodfellow2014generative}
I.~Goodfellow, J.~Pouget-Abadie, M.~Mirza, B.~Xu, D.~Warde-Farley, S.~Ozair,
  A.~Courville, and Y.~Bengio.
\newblock Generative {A}dversarial {N}ets.
\newblock In {\em Advances in Neural Information Processing Systems}, pages
  2672--2680, 2014.

\bibitem{im2015denoising}
D.~J. Im, S.~Ahn, R.~Memisevic, and Y.~Bengio.
\newblock Denoising criterion for variational auto-encoding framework.
\newblock {\em arXiv preprint arXiv:1511.06406}, 2015.

\bibitem{kasmi2016classification}
R.~Kasmi and K.~Mokrani.
\newblock Classification of malignant melanoma and benign skin lesions:
  implementation of automatic abcd rule.
\newblock {\em IET Image Processing}, 10(6):448--455, 2016.

\bibitem{kingma2013auto}
D.~P. Kingma and M.~Welling.
\newblock Auto-encoding variational {B}ayes.
\newblock In {\em Proceedings of the 2015 International Conference on Learning
  Representations (ICLR-2015), arXiv preprint arXiv:1312.6114}, 2014.

\bibitem{li2017skin}
Y.~Li and L.~Shen.
\newblock Skin lesion analysis towards melanoma detection using deep learning
  network.
\newblock {\em arXiv preprint arXiv:1703.00577}, 2017.

\bibitem{makhzani2015adversarial}
A.~Makhzani, J.~Shlens, N.~Jaitly, and I.~Goodfellow.
\newblock Adversarial autoencoders.
\newblock {\em arXiv preprint arXiv:1511.05644}, 2015.

\bibitem{ISIC-Menegola17}
A.~Menegola, J.~Tavares, M.~Fornaciali, L.~T. Li, S.~E.~F. de~Avila, and
  E.~Valle.
\newblock {RECOD} titans at {ISIC} challenge 2017.
\newblock {\em CoRR}, abs/1703.04819, 2017.

\bibitem{radford2015unsupervised}
A.~Radford, L.~Metz, and S.~Chintala.
\newblock Unsupervised representation learning with deep convolutional
  generative adversarial networks.
\newblock In {\em International Conference on Learning Representations (ICLR)
  2016, arXiv preprint arXiv:1511.06434}, 2015.

\bibitem{ramlakhan2011mobile}
K.~Ramlakhan and Y.~Shang.
\newblock A mobile automated skin lesion classification system.
\newblock In {\em Tools with Artificial Intelligence (ICTAI), 2011 23rd IEEE
  International Conference on}, pages 138--141. IEEE, 2011.

\bibitem{round2000lesion}
A.~J. Round, A.~W. Duller, and P.~J. Fish.
\newblock Lesion classification using skin patterning.
\newblock {\em Skin research and technology}, 6(4):183--192, 2000.

\bibitem{schmid1997colour}
P.~Schmid and S.~Fischer.
\newblock Colour segmentation for the analysis of pigmented skin lesions.
\newblock In {\em Image Processing and Its Applications, 1997., Sixth
  International Conference on}, volume~2, pages 688--692. IET, 1997.

\bibitem{vincent2008extracting}
P.~Vincent, H.~Larochelle, Y.~Bengio, and P.-A. Manzagol.
\newblock Extracting and composing robust features with denoising autoencoders.
\newblock In {\em Proceedings of the 25th {I}nternational {C}onference on
  {M}achine {L}earning}, pages 1096--1103. ACM, 2008.

\bibitem{vincent2010stacked}
P.~Vincent, H.~Larochelle, I.~Lajoie, Y.~Bengio, and P.-A. Manzagol.
\newblock Stacked denoising autoencoders: Learning useful representations in a
  deep network with a local denoising criterion.
\newblock {\em Journal of Machine Learning Research}, 11(Dec):3371--3408, 2010.

\end{thebibliography}

\pagebreak

\onecolumn
\section{Appendices}
\label{Appendices}
Table. \ref{CNN_architecture}) shows details of the architecture of the CNN baseline.

\centering
\begin{table}[h]
\centering
\caption{CNN architecture}
\label{CNN_architecture}
\begin{tabular}{|c|c|}
\hline
\textbf{Input}                        & \textbf{Real Image}                                              \\ \hline
\multirow{11}{*}{\textbf{Layers}}     & \textbf{conv2D} {[} filterSize : 5, nFilters : 64, stride=2, padding=2{]}  \\
                                      & \textbf{Relu}                                                    \\
                                      & \textbf{conv2D} {[} filterSize : 5, nFilters : 128, stride=2, padding=2{]}  \\
                                      & \textbf{Relu}                                                    \\
                                      & \textbf{conv2D} {[} filterSize : 5, nFilters : 256, stride=2, padding=2{]}  \\
                                      & \textbf{Relu}                                                    \\
                                      & \textbf{conv2D} {[} filterSize : 5, nFilters : 512, stride=2, padding=2{]} \\
                                      & \textbf{Relu}                                                    \\
                                      & \textbf{Linear}  {[}Size :  1000{]}                                       \\
                                      & \textbf{Linear}  {[}Size :  1{]}                                          \\
                                      & \textbf{Sigmoid}                                                 \\ \hline
\multicolumn{1}{|l|}{\textbf{Output}} & Probability of label = 1                                         \\ \hline
\end{tabular}
\end{table}

The tables bellow show details of the architecture for the ssDAAE. Table \ref{Encoder} shows the encoder and Table \ref{Decoder} shows the decoder. In addition, the label discriminator is shown in Table \ref{disY} and the latent discriminator is shown in Table \ref{disZ}.

\begin{table}[h]
\centering
\caption{Encoder}
\label{Encoder}
\begin{tabular}{|c|cc|}
\hline
Input                             & \multicolumn{2}{c|}{\textbf{Real Image}}                                                    \\ \hline
\multirow{11}{*}{\textbf{Layers}} & \multicolumn{2}{c|}{\textbf{conv2D} {[} filterSize : 5, nFilters : 64, stride=2, padding=2{]}}         \\
                                  & \multicolumn{2}{c|}{\textbf{Relu}}                                                                    \\
                                  & \multicolumn{2}{c|}{\textbf{conv2D} {[} filterSize : 5, nFilters : 128, stride=2, padding=2{]}}         \\
                                  & \multicolumn{2}{c|}{\textbf{Relu}}                                                                    \\
                                  & \multicolumn{2}{c|}{\textbf{conv2D} {[} filterSize : 5, nFilters : 256, stride=2, padding=2{]}}         \\
                                  & \multicolumn{2}{c|}{\textbf{Relu}}                                                                    \\
                                  & \multicolumn{2}{c|}{\textbf{conv2D} {[} filterSize : 5, nFilters : 512, stride=2, padding=2{]}}        \\
                                  & \multicolumn{2}{c|}{\textbf{Relu}}                                                                    \\
                                  & \multicolumn{2}{c|}{\textbf{Linear}  {[}Size :  1000{]}}                                              \\
                                  & \multicolumn{1}{c|}{\textbf{Linear}  {[}Size :  1{]}} & \multirow{2}{*}{\textbf{Linear}  {[}Size :  200{]}}   \\
                                  & \multicolumn{1}{c|}{\textbf{Sigmoid}}                 &                                              \\ \hline
Encoder Output                    & \multicolumn{1}{c|}{\textbf{Label}}          & \multicolumn{1}{c|}{\textbf{Encoded Vector}} \\ \hline
\end{tabular}
\end{table}

\begin{table}[h]
\centering
\caption{Decoder. Conv2D$^T$ represents transposed 2D convolutions.}
\label{Decoder}
\centering
\begin{tabular}{|c|cc|}
\hline
\textbf{Input}                    & \multicolumn{1}{c|}{\textbf{Label}}                & \multicolumn{1}{c|}{\textbf{Encoded Vector}}                \\ \hline
\multirow{11}{*}{\textbf{Layers}} & \multicolumn{2}{c|}{\textbf{Concat}}                                                                              \\
                                  & \multicolumn{2}{c|}{\textbf{Linear}  {[}Size :  512x(4x4){]}}                                                                    \\
                                  & \multicolumn{2}{c|}{\textbf{Relu}}                                                                                \\
                                  & \multicolumn{2}{c|}{\textbf{Conv2D$^T$} {[}filterSize = 3, nFilters : 256, stride=2, padding=1, output\_padding=1{]}} \\
                                  & \multicolumn{2}{c|}{\textbf{Relu}}                                                                                \\
                                  & \multicolumn{2}{c|}{\textbf{Conv2D$^T$} {[}filterSize = 3, nFilters : 128, stride=2, padding=1, output\_padding=1{]}} \\
                                  & \multicolumn{2}{c|}{\textbf{Relu}}                                                                                \\
                                  & \multicolumn{2}{c|}{\textbf{Conv2D$^T$} {[}filterSize = 3, nFilters : 64, stride=2, padding=1, output\_padding=1{]}} \\
                                  & \multicolumn{2}{c|}{\textbf{Relu}}                                                                                \\
                                  & \multicolumn{2}{c|}{\textbf{Conv2D$^T$} {[}filterSize = 3, nFilters : 3, stride=2, padding=1, output\_padding=1{]}} \\
                                  & \multicolumn{2}{c|}{\textbf{Sigmoid}}                                                                             \\ \hline
\textbf{Decoder Output}           & \multicolumn{2}{c|}{Reconstruction Image}                                                                        \\ \hline
\end{tabular}
\end{table}

\begin{table}[h]
\centering
\caption{Regularisation of the classifier}
\label{disY}
\begin{tabular}{|c|c|}
\hline
\textbf{Input}                   & \textbf{Label}                  \\ \hline
\multirow{4}{*}{\textbf{Layers}} & \textbf{Linear}  {[}Size :  1000{]}      \\
                                 & \textbf{Relu}                   \\
                                 & \textbf{Linear}  {[}Size :  1{]}         \\
                                 & \textbf{Sigmoid}                \\ \hline
\textbf{Discriminator Output}          & Label Discriminator Probability \\ \hline
\end{tabular}
\end{table}

\begin{table}[h]
\centering
\caption{Regularisation of the encoder}
\label{disZ}
\begin{tabular}{|c|c|}
\hline
\textbf{Input}                   & \textbf{Encoded Vector}                  \\ \hline
\multirow{4}{*}{\textbf{Layers}} & \textbf{Linear}  {[}Size :  1000{]}      \\
                                 & \textbf{Relu}                   \\
                                 & \textbf{Linear}  {[}Size :  1{]}         \\
                                 & \textbf{Sigmoid}                \\ \hline
\textbf{Discriminator Output}          & Latent Space Discriminator Probability \\ \hline
\end{tabular}
\end{table}

\end{document}